\pdfoutput=1

\documentclass[pageno]{jpaper}

\newcommand\behzad[1]{\textcolor{black}{#1}}
\newcommand\fff[1]{\textcolor{black}{#1}}
\newcommand\onur[1]{\textcolor{black}{#1}}

\usepackage{dblfloatfix} 
\usepackage{geometry}%
\usepackage{lipsum}
\usepackage{multicol}%
\usepackage{wasysym}
\usepackage{amssymb}
\usepackage{fancyhdr}
\usepackage[normalem]{ulem}
\usepackage{hyperref}
\usepackage{graphicx}
\usepackage{caption}
\usepackage{authblk}
\usepackage{textcomp}

\makeatletter
\renewcommand\AB@affilsepx{\hspace{30pt}\protect\Affilfont}

\makeatother

\begin{document}

\title{\textcolor{black}{An Experimental Study of Reduced-Voltage Operation in Modern FPGAs\\for Neural Network Acceleration}}

\author[1]{Behzad Salami}
\author[2]{Erhan Baturay Onural}
\author[2]{Ismail Emir Yuksel}
\author[2]{Fahrettin Koc}
\author[2]{Oguz Ergin}
\author[1,3]{Adri\'{a}n Cristal Kestelman}
\author[1]{Osman S. Unsal}
\author[4]{Hamid Sarbazi-Azad}
\author[5]{Onur Mutlu}
\affil[1]{\textit{BSC}}
\affil[2]{\textit{TOBB ET\"{U}}}
\affil[3]{\textit{UPC and CSIC-IIIA}}
\affil[4]{\textit{SUT and IPM}}
\affil[5]{\textit{ETH Z\"{u}rich}}


\date{}
\maketitle

\thispagestyle{empty}

\begin{abstract}
\behzad{We} empirically evaluate an undervolting technique, \fff{i.e., underscaling the circuit supply voltage} below the \behzad{nominal} level\behzad{,} to improve the power-efficiency of Convolutional Neural \behzad{Network (CNN) accelerators mapped to} Field Programmable Gate Arrays (FPGAs). \behzad{Undervolting} below \behzad{a} safe \behzad{voltage} level can lead to timing faults due to \behzad{excessive} circuit latency \fff{increase}. \behzad{We} evaluate the reliability-power trade-off for such accelerators. Specifically, we experimentally study the \behzad{reduced-voltage operation} of multiple components of real FPGAs, characterize the \behzad{corresponding} reliability behavior of CNN accelerators, propose techniques to minimize the \behzad{drawbacks} of reduced-voltage \behzad{operation}, and combine undervolting with architectural CNN optimization techniques, \fff{\textit{i.e.,}} quantization and pruning. \behzad{We} investigate the effect of \behzad{environmental} temperature on the reliability-power trade-off of such accelerators. 

We perform experiments on three identical samples of modern Xilinx ZCU102 \behzad{FPGA} platforms with five state-of-the-art image classification CNN benchmarks. This approach allows us to study the effects of \behzad{our} undervolting technique for both software and hardware variability. We \behzad{achieve more} than 3X power-efficiency ($GOPs/W$) gain \behzad{via} undervolting. 2.6X of this gain is the result of eliminating the voltage guardband region, \behzad{\textit{i.e.,} the} safe voltage region below the nominal level that is set by FPGA vendor to ensure \behzad{correct} functionality \behzad{in} worst-case environmental and \onur{circuit} \behzad{conditions}. \behzad{43\% of the} power-efficiency gain is due \fff{to} further undervolting below the guardband\behzad{, which} comes at the cost of accuracy loss in the CNN accelerator. \fff{We evaluate an effective frequency underscaling technique that prevents this accuracy loss, and find that it reduces the power-efficiency gain from 43\% to 25\%.}

\end{abstract}

\section{Introduction}
\label{sec:introduction}
Deep Neural Networks (DNNs) and specifically Convolutional Neural Networks (CNNs) have recently attained significant success in image and video classification tasks. \behzad{They} are fundamental for state-of-the-art real-world applications running on embedded systems as well as data centers. \behzad{These neural networks learn a model} from a dataset in \behzad{their} training phase and make predictions on \behzad{new, previously-unseen} data in \behzad{their} classification phase. 
However, their power-efficiency is inherently the primary concern due to the massive amount of data movement and computational power required. Thus, the scalability of CNNs for enterprise applications and \behzad{their deployment} in battery-limited scenarios, such as \behzad{in} drones and mobile devices, \behzad{are crucial concerns}.

Typically, hardware acceleration using Graphics Processing Units (GPUs)~\cite{zhang2018shufflenet}, \behzad{Field Programmable Gate Arrays (FPGAs)~\cite{sharma2016high,qiu2016going}}, or Application-Specific Integrated Circuits (ASICs)~\cite{sharma2018bit,jouppi2017datacenter,chen2016eyeriss} leads to a significant reduction \behzad{in CNN} power consumption~\cite{sze2017efficient}. Among \behzad{these}, FPGAs are rapidly becoming popular and are expected to be used in 33\% of modern data centers by 2020 \cite{top500}. This increase in the popularity of FPGAs is attributed to their power-efficiency compared to GPUs, their flexibility compared to ASICs, and \behzad{recent} advances in High-Level Synthesis (HLS) tools that significantly facilitate easier \behzad{mapping of applications on FPGAs}~\cite{putnam2014reconfigurable,vaishnav2018survey, salami2015hatch, salami2016accelerating,salami2017axledb,park2017scale,arcas2016hardware}. Hence, major companies, such as Amazon \cite{karandikar2018firesim} \behzad{(with EC2 F1 cloud)} and Microsoft~\cite{fowers2018configurable} \behzad{(with Brainwave project)}, have made large investments in FPGA-based CNN accelerators. However, recent studies show that FPGA-based accelerators are at least 10X less power-efficient compared to ASIC-based ones~\cite{boutros2018you,nurvitadhi2016accelerating,nurvitadhi2019compete}. In this paper, we aim to bridge this power-efficiency gap by \behzad{empirically understanding and} leveraging an effective undervolting technique for FPGA-based CNN accelerators. 



\behzad{Power-efficiency} of state-of-the-art CNNs \behzad{generally} improves \behzad{via} architectural-level techniques, such as \behzad{quantization~\cite{zhu2019configurable} and pruning~\cite{molchanov2016pruning}}. These techniques do not \behzad{significantly} compromise \behzad{CNN} accuracy as they \behzad{exploit} the sparse nature of \behzad{CNN} applications~\cite{parashar2017scnn, albericio2016cnvlutin,zhang2016cambricon}. To further improve the power-efficiency of FPGA-based CNN accelerators, we propose to employ an orthogonal \behzad{hardware}-level approach\behzad{: undervolting (\textit{i.e.,} circuit supply voltage underscaling)} below the nominal/default level ($V_{nom}$), combined with the aforementioned architectural-level techniques. \behzad{FPGA vendors usually add a} voltage guardband to ensure the correct operation of FPGAs under the worst-case \behzad{circuit} and environmental conditions. However, these guardbands can be very conservative and unnecessary for state-of-the-art applications. Supply voltage underscaling below the nominal level \behzad{was} already shown \behzad{to provide} significant efficiency \behzad{improvements in} CPUs~\cite{papadimitriou2019adaptive}, GPUs~\cite{zou2018voltage,miller2012vrsync}, ASICs~\cite{chandramoorthy2019resilient}, and \behzad{DRAMs~\cite{chang2017understanding,koppula2019eden}}. This paper extends such studies \behzad{to FPGAs. Specifically, we study the classification phase of FPGA-based CNN accelerators}, as this phase can be repeatedly used in power-limited edge devices (unlike the training phase, which is \behzad{invoked much less frequently}). \behzad{Unlike} simulation-based approaches that may not be \fff{accurate enough}~\cite{zhang2018thundervolt, salamat2019workload}, our study is based on real off-the-shelf FPGA devices. 

The extra voltage guardband can range between 12-35\% of the nominal supply voltage of modern CPUs~\cite{papadimitriou2019adaptive}, GPUs~\cite{zou2018voltage}, and DRAM \behzad{chips}~\cite{chang2017understanding}. Reducing the supply voltage in this guardband region does \behzad{\emph{not}} lead to \behzad{reliability issues under normal operating conditions,} and thus, eliminating this guardband \behzad{can} result in a significant power reduction for \behzad{a wide variety of} real-world applications. \textcolor{black}{We experimentally \behzad{demonstrate} a large voltage guardband for modern FPGAs\behzad{: an} average of 33\% with a slight variation across hardware platforms and software benchmarks. Eliminating this guardband leads to significant power-efficiency \behzad{($GOPs/W$) improvement}, on average, 2.6X, without any performance or reliability overheads. 
 With further undervolting, the power\behzad{-efficiency improves} by an extra 43\%, leading to a total \behzad{improvement} of more than 3X. This additional \behzad{gain} does not come for free, as we observe \behzad{exponentially}-increasing CNN accuracy loss below the guardband region. With further undervolting below this guardband, our experiments indicate that the \behzad{minimum} supply voltage at which the internal FPGA components could be functional ($V_{crash}$) is equal to, on average, 63\% of $V_{nom}$. Further reducing the supply voltage results in system crash.}

\textcolor{black}{We evaluate our undervolting technique on three identical samples of the Zynq-based ZCU102 platform \cite{zcu102}, a representative modern FPGA from Xilinx. However, we believe that our experimental observations are applicable to other FPGA platforms \behzad{as well}, perhaps with some minor differences. We previously showed \behzad{benefits of reduced-voltage operation} for on-chip memories on \behzad{different, }older FPGA \behzad{platforms}~\cite{salami2018comprehensive}. \behzad{Other works observed similar behavior} for different \behzad{types} of CPUs~\cite{papadimitriou2019adaptive}, GPUs~\cite{zou2018voltage}, and DRAM \behzad{chips}~\cite{chang2017understanding}. In this paper, we characterize the power dissipation of \behzad{FPGA-based CNN accelerators under reduced-voltage levels and} apply undervolting \behzad{to} improve the power-efficiency of such \behzad{accelerators}.}\footnote{ \textcolor{black}{\behzad{Our} exploration of the FPGA voltage behavior and the subsequent power-efficiency gain is applicable to any application.}} 

\behzad{We experimentally evaluate the effects of reduced-voltage operation in} on-chip components of the FPGA platform, including Block RAMs (BRAMs) and internal FPGA \behzad{components,} containing Look-Up Tables (LUTs), Digital Signal Processors (DSPs), buffers, and routing resources.\footnote{These internal FPGA components share a single voltage rail in the studied FPGA platform. To our knowledge, \behzad{such voltage rail sharing} is a typical case for most modern FPGA platforms.} We \behzad{perform} our experiments on five state-of-the-art CNN image classification benchmarks, including VGGNet~\cite{miniV}, GoogleNet~\cite{miniG}, AlexNet~\cite{alex}, ResNet~\cite{he2016deep}, and Inception~\cite{miniG}. This enables us to experimentally study the workload-to-workload variation on the power-reliability trade-offs of FPGA-based CNN \behzad{accelerators. Specifically}, we extensively characterize the reliability behavior of the studied benchmarks below the guardband level and evaluate a frequency underscaling technique to prevent the accuracy loss in this voltage region. Our study also \behzad{examines} the effects of architectural quantization and pruning techniques with reduced-voltage FPGA operation. Finally, we experimentally evaluate the \behzad{effect} of environmental temperature variation on the power-reliability behavior of \behzad{FPGA-based CNN accelerators}. 


\subsection{Contributions}
To our knowledge, for the first time, this paper experimentally studies the power-performance-accuracy characteristics of CNN accelerators with greatly reduced supply voltage capability implemented in real FPGAs. In summary, we achieve a total of more than 3X power-efficiency improvement for FPGA-based CNN accelerators. \behzad{We} gain insights into the reduced-voltage operation of such accelerators and, in turn, the \behzad{effect} of FPGA supply voltage on the power-reliability trade-off. \behzad{We make the following major contributions:}
\begin{itemize}
\item We characterize the power consumption of FPGA-based CNN accelerators across \behzad{different FPGA components}. We identify that the internal on-chip components, including processing elements, \behzad{contribute to} a vast majority of the total power consumption. \behzad{We reduce} this source of power consumption \behzad{via} our undervolting technique. 

\item \textcolor{black}{We improve the power-efficiency of FPGA-based CNN accelerators \behzad{by} more than 3X\behzad{, measured across five state-of-the-art image classification benchmarks}. 2.6X of the power-efficiency gain is due to eliminating the voltage guardband\behzad{, which we measure} to be on average 33\%. An additional 43\% gain is due to further undervolting below the guardband\behzad{, which} comes at the cost of CNN accuracy loss.} 

\item \behzad{We} characterize the reliability behavior of \behzad{FPGA}-based CNN accelerators when executed below the voltage guardband level and observe an exponential reduction \behzad{in} CNN accuracy \behzad{as voltage reduces}. We observe that workloads with more parameters, \textit{e.g.,} ResNet and Inception, are relatively more vulnerable to undervolting-related faults.  

\item To prevent \behzad{CNN} accuracy loss below the voltage guardband level, we combine voltage underscaling with frequency underscaling. We experiment with a supply voltage lower than $V_{nom}$ and with operating frequency $F_{op} < F_{max}$. Our experiments \behzad{show} that the most \behzad{\emph{energy-efficient}} operating point is the one with the maximum frequency and minimum safe voltage, namely, $V_{min}$. However, lower voltage and lower frequency lead to better \behzad{\emph{power-efficiency}}.

\item We combine voltage underscaling with \behzad{the} existing CNN quantization and pruning techniques and study the power-reliability trade-off of \behzad{such} optimized FPGA-based CNN \behzad{accelerators}. We observe that these bit/parameter-size reduction techniques \behzad{(quantization and pruning)} slightly increase the vulnerability of a CNN to undervolting-related faults; but, they deliver significantly higher power-efficiency when integrated with our undervolting technique.

\item We study the effect of environmental temperature on the power-reliability trade-off of reduced-voltage FPGA-based CNN accelerators. \behzad{We} observe that temperature has a direct \behzad{effect on} the power consumption of such accelerators. However, at \behzad{very} low voltage levels, this \behzad{effect} is not noticeable. 

\item We evaluate the effect of hardware platform variability by repeating \behzad{our} experiments on three identical samples of the Xilinx ZCU102 \behzad{FPGA} platform. We \behzad{find} large voltage \behzad{guardbands} in all platforms (an average of 33\%), \textit{i.e.,} $V_{min}= 0.67 * V_{nom}= 570mV$. \behzad{However, across three FPGAs, we observe a variation on $V_{min}$\textit{, i.e.,} $\Delta V_{min}=31mV$}. This variation can be due to process variation. \fff{Our results show that} the variation of \behzad{guardband} regions \behzad{across} different CNN workloads is insignificant.   

\end{itemize}


\section{Background}
\label{sec:background}
In this section, we briefly introduce the most important concepts used in this paper, including the architecture of CNNs as well as the undervolting technique.

\subsection{Convolutional Neural Networks (CNNs)}
DNNs are a class of Machine Learning (ML) methods that are designed to classify unseen objects or entities using non-linear transformations applied to \behzad{input data} \cite{lecun2015deep}. DNNs are composed of biologically inspired neurons, interconnected to each other. Among different DNN models, multi-layer CNNs are a common type, which has recently shown acceptable success in \behzad{classification} tasks for real-world applications.  

\subsubsection{Phases of a CNN: Training and Classification.}
\behzad{A CNN model encompasses} two \behzad{stages:} training and classification (inference). Training \behzad{learns a model from a set of training data. It} is an iterative, usually \behzad{a} single-time \behzad{(or relatively infrequently-executed) step, including} backward and forward phases. It \behzad{adjusts} the CNNs parameters, \textit{i.e.,} weights and biases, which determine the strength of the \behzad{connections} between different neurons \behzad{across} CNN layers. The training phase \behzad{minimizes} a loss function, which directly relates to \behzad{the accuracy of the neural network in the classification phase}. \behzad{In contrast,} inference is a post-training phase \behzad{that} aims to classify \behzad{unknown data, using the trained network model}. The inference phase is more \behzad{frequently executed in} edge devices with power-\behzad{constrained} environments. \behzad{The} target of this paper \behzad{is the} inference stage, similar to \behzad{many} existing efforts on the acceleration \behzad{of CNNs~\cite{guo2017survey,sze2017efficient, koppula2019eden}}. 

\subsubsection{Internal Architecture of a CNN.}
\behzad{A CNN is} composed of multiple processing layers such as Convolution, Pooling, Fully-Connected, and SoftMax for feature extraction with various abstractions. Other customized layers can be used case by case for more optimized feature extraction, such as Batch \behzad{Normalization~\cite{nakahara2017batch}}. The functionality of each type of layer depends on the way \behzad{in which} the neurons are interconnected. Convolution layers generate a more profound abstraction of the input data, called a feature map. Following each Convolution layer, there is usually a Max/Avg Pooling layer to reduce the \behzad{dimensionality} of the feature map. Successive multiple Convolution and Pooling layers generate in-depth information from the input data. Afterward, Fully-Connected layers are typically applied for classification purposes. Finally, the SoftMax layer generates the class probabilities from the class scores in the output layer. Between layers, there are activation functions, such as Relu or Sigmoid, to \behzad{add} non-linear \fff{properties} to the network. \behzad{The} required computations of different layers are translated to matrix multiplication computations. Thus, \behzad{matrix} multiplication optimization techniques, such as FFT \behzad{or Strassen~\cite{lavin2016fast}}, can be applied to accelerate the \behzad{inference} implementation. Matrix multiplication is an ideal application to take advantage of parallel and data flow execution model \behzad{used in FPGA-based} hardware accelerators.


\subsubsection{Architectural Optimizations.}
To improve the power-efficiency of CNNs\behzad{, two} most \behzad{commonly-used} architectural-level techniques are quantization~\cite{zhou2017incremental} and pruning~\cite{molchanov2016pruning}.\footnote{\behzad{There are also other techniques, such as batching~\cite{shen2017escher}, loop unrolling~\cite{ma2017optimizing}, and memory compression~\cite{kim2015compression}.}} These \behzad{two} techniques rely on the \behzad{sparse nature} of CNNs, \behzad{\textit{i.e.,}} a vast majority of CNN computations are unnecessary. Quantization aims to reduce the complexity of \behzad{high-precision CNN} computation units by \behzad{substituting selected} floating-point parameters \behzad{with} low-precision fixed-point. \behzad{Pruning} aims to reduce the model size by eliminating \behzad{unnecessary} weight/neurons/connections of \behzad{a} CNN. These \fff{architectural} techniques are applicable to any underlying hardware. There are numerous extensions of quantization~\cite{zhou2017incremental,zhu2019configurable} and pruning~\cite{yazdani2018dark,he2017channel} techniques. In our experiments, we integrate typical \behzad{quantization~\cite{han2016eie} and pruning~\cite{han2015learning}} techniques with \behzad{our} proposed \behzad{hardware}-level undervolting technique to \behzad{further} improve the power-efficiency of FPGA-based CNN accelerators.

\subsection{Undervolting: Supply Voltage Underscaling Below the Nominal Voltage Level}
The total power consumption of any hardware \fff{substrate} is directly related to its supply voltage\behzad{: quadratically} and linearly with \behzad{dynamic} and static power, respectively. Thus, supply voltage underscaling toward the threshold voltage significantly reduces power consumption. Voltage underscaling is a common power-saving approach \behzad{as} manufacturing technology node \behzad{size reduces}. For instance, the $V_{nom}$ of Xilinx FPGAs is $1V$, $0.9V$, and $0.85V$ for \behzad{implementations in} 28nm, 20nm, and 16nm \behzad{technology nodes}, respectively. The aim of our undervolting technique \behzad{is to reduce the supply voltage below} the default $V_{nom}$. 
However, circuit latency can increase substantially \behzad{ when supply voltage is reduced} below the guardband level, and in turn, timing faults can appear. These timing faults are manifested as bit-flips in memories or logic timing violations in \behzad{data} paths. They can potentially \behzad{cause} the application to produce wrong results, \behzad{leading to} reduced accuracy in CNNs\behzad{, or}, in the worst-case, they may cause system crashes. \behzad{There} are several approaches \behzad{to deal with undervolting faults}, such as preventing these faults by\behzad{: \textit{i)}} simultaneously decreasing the frequency~\cite{tang2019impact}\behzad{, which has an associated} performance degradation \behzad{cost}, \fff{\textit{ii)}} \behzad{fixing} the faults by using fault mitigation techniques, such as Error Correction Codes (ECCs) for memories~\cite{bacha2014using,salami2019evaluating} and Razor shadow latches for data paths~\cite{ernst2003razor}\behzad{, which comes at} the cost of extra hardware, or \behzad{\textit{iii)}} architectural improvements, such as additional iterations in CNN \behzad{training~\cite{zhang2018analyzing} that may incur \fff{hardware and/or software} adaptation costs}. 

\behzad{There} are two approaches \behzad{to} undervolting studies\behzad{: \textit{i)}} simulation-based \behzad{studies~\cite{roelke2017pre,yalcin2016exploring,zhang2018thundervolt, swaminathan2017bravo}}, or \behzad{\textit{ii)}} direct implementation \behzad{or testing} on real hardware fabrics, mainly performed on CPUs, GPUs, ASICs, and \behzad{DRAMs~\cite{zou2018voltage,bacha2014using,papadimitriou2019adaptive,  chang2017understanding, 8416495,koppula2019eden}}. The simulation-based approach requires less engineering \behzad{effort}. However, \fff{validation of} \behzad{simulation} results on real hardware is the primary concern \behzad{with such an approach}. \behzad{In contrast}, \behzad{the real hardware} \fff{evaluation} approach \fff{requires} substantial engineering \behzad{effort}, and it \behzad{is device-} and vendor-dependent. Such \behzad{a real hardware approach} leads to exact experimental results \behzad{and} it provides an opportunity to \behzad{study} device-dependent parameters, such as voltage guardbands and real power and reliability behavior of underlying hardware. In this paper, we follow the \behzad{real hardware} approach by evaluating \behzad{undervolting} on real modern off-the-shelf FPGA devices for state-of-the-art CNN workloads and benchmarks. 

\section{Experimental Methodology}
\label{sec:methodology}
\begin{figure}[H]
\centering
\includegraphics[width=\linewidth]{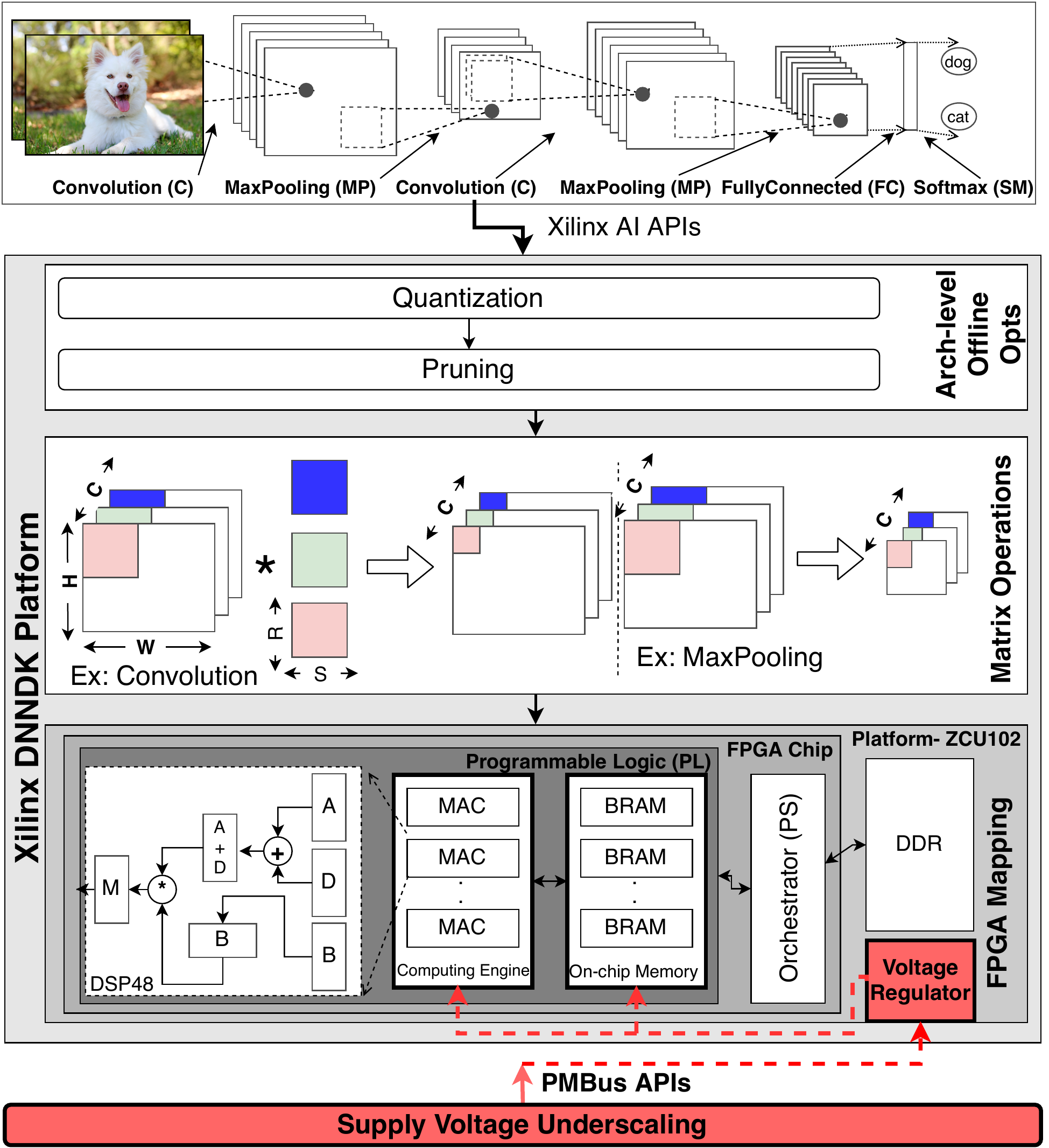}
\caption{Our overall methodology (for simplicity, we show a simplified block diagram of Xilinx DNNDK}
\label{fig:overall} 
\end{figure}

Figure~\ref{fig:overall} \behzad{depicts} the overall \behzad{methodological} flow of our experiments. In this section, we elaborate on its different aspects, including \behzad{our} implementation methodology, benchmarks, and \behzad{undervolting} methodology of \behzad{our} FPGA platform.

\begin{table*}[]
\caption{Evaluated CNN Benchmarks.}
\label{table:benchmarks}
\begin{tabular}{|l||l|r|r||r|r||r|r|}
\hline
\multicolumn{1}{|c||}{\textbf{CNN}} & \multicolumn{3}{c||}{\textbf{Dataset}} & \multicolumn{2}{c||}{\textbf{Parameters}} & \multicolumn{2}{c|}{\textbf{Inference Accuracy (\%)}} \\ \cline{2-8} 
\multicolumn{1}{|c||}{\textbf{Model}} & \textbf{Name} & \multicolumn{1}{l|}{\textbf{Inputs}} & \multicolumn{1}{l||}{\textbf{Outputs}} & \multicolumn{1}{l|}{\textbf{\#Layers}} & \multicolumn{1}{l||}{\textbf{Size}} & \multicolumn{1}{l|}{\textbf{Literature}} & \multicolumn{1}{l|}{\textbf{Our \behzad{design} @Vnom}} \\ \hline \hline
\textbf{VGGNet} & Cifar-10 & 32*32 & 10 &  6& 8.7MB & 87\% \cite{miniV} & 86\% \\ \hline
\textbf{GoogleNet} & Cifar-10 & 32*32 & 10 & 21 & 6.6MB & 91\% \cite{miniG} & 91\% \\ \hline
\textbf{AlexNet} & Kaggle Dogs vs. Cats & 227*227 & 2 & 8 & 233.2MB & 96\% \cite{alex} & 92.5\% \\ \hline
\textbf{ResNet50} & ILSVRC2012 & 224*224 & 1000 &  50& 102.5MB & 76\% \cite{he2016deep} & 68.8\% \\ \hline
\textbf{Inception} & ILSVRC2012 & 224*224 & 1000 &  22& 107.3MB & 68.7\% \cite{miniG} & 65.1\% \\ \hline
\end{tabular}
\end{table*} 

\subsection{CNN Model Development Platform}
\textcolor{black}{For \behzad{our} implementation, we leverage \fff{the} Deep Neural Network Development Kit (DNNDK)~\cite{dnndk}, a CNN framework from Xilinx. DNNDK is an integrated framework to facilitate CNN development and deployment on Deep learning \behzad{Processing} Units (DPUs). \behzad{In} this paper, we use DNNDK as \behzad{it is} a freely-available framework instead of a specialized custom design, to ensure that the results reported in this paper are reproducible and general-enough for state-of-the-art CNN implementations. Although we do not expect a significant difference by experimenting on DNNDK versus other DNN platforms, our future plan is to verify this by repeating the experiments on other platforms, such as DNNWeaver~\cite{sharma2016high}.} DNNDK provides a complete set of toolchains with compression, compilation, deployment, and profiling, for the mapping of CNN classification phases onto FPGAs integrated with hard CPU cores via a comprehensive and easy-to-use C/C++ programming interface. 

Among the components of DNNDK, the DEep ComprEssioN Tool (DECENT) is responsible for quantization and pruning tasks. The quantization utility of \behzad{DECENT} can convert \behzad{a \fff{floating-point} CNN} model to a quantized model with the precision of at most INT8~\cite{han2016eie}. \behzad{The} pruning utility aims to minimize the model size by removing unnecessary connections of the CNN~
\cite{han2015learning}. \behzad{We} perform our baseline evaluation on a model with INT8 precision and without any pruning optimization. However, in Section~\ref{subsec:architectural}, \behzad{we evaluate different configurations to provide} a more comprehensive analysis. 
There are different sizes of soft DPUs provided by DNNDK with various hardware utilization \behzad{rates}~\cite{DPU}. Among them, B4096 is the \behzad{largest} model that utilizes a maximum \behzad{fraction} of BRAMs and DSPs, \textit{i.e.,} 24.3\% and 25.6\%, respectively, \behzad{resulting} in a peak performance of 4096 operations/cycle with a default DPU frequency of 333Mhz and DSP frequency of 666Mhz. In total, a maximum of three B4096 DPUs can be \behzad{used} in the hardware platform evaluated in this paper. Our experiments are based on the B4096 configuration to achieve peak performance.

\subsection{CNN Benchmarks}
We evaluate undervolting in FPGA-based CNN accelerators with \behzad{five} commonly-used image classification \behzad{benchmarks, shown in Table \ref{table:benchmarks}:} VGGNet \cite{miniV}, GoogleNet \cite{miniG}, AlexNet \cite{alex}, ResNet \cite{he2016deep}, and Inception \cite{miniG}. \behzad{To perform} a comprehensive analysis and \fff{study} workload-to-workload variation \fff{better,} we choose models \behzad{whose parameter sizes vary from} a few MBs, \textit{e.g.,} GoogleNet, to hundreds of MBs, \textit{e.g.,} AlexNet. \behzad{Our} benchmarks have \behzad{different numbers and types} of layers, as \behzad{shown} in Table~\ref{table:benchmarks}. \behzad{The} default activation function used \behzad{in} benchmarks is Relu.

\subsection{Undervolting}
In this section, we briefly explain the prototype FPGA platform and the associated voltage \behzad{control} setup. 

\subsubsection{Prototype FPGA Platform.}
Our prototype is based on the Xilinx ZCU102 \behzad{FPGA} platform fabricated at a 16nm technology node. We choose this platform because \behzad{it} is \behzad{\textit{i)}} equipped \behzad{with} voltage underscaling capability\behzad{, \textit{ii)}} supported by DNNDK. We repeat experiments on three identical samples of ZCU102 to \behzad{study} the effect of \behzad{hardware} platform variability. ZCU102 is populated with the Zynq UltraScale+ XCZU9EG-2FFVB1156E MPSoC that combines a Processing System (PS) and user-Programmable Logic (PL) \behzad{in} the same device. The PS part features a quad-core \behzad{64-bit} \fff{ARM} Cortex-A53 and is mainly used for the host communication in DNNDK. The PL \behzad{part} has 32.1Mbit of BRAMs, 600K LUTs, and 2520 DSPs. For the CNN implementation, DPUs are mapped into the PL \behzad{side}. As mentioned earlier, \behzad{our} baseline hardware configuration employs three B4096 DPUs, the maximum possible number, leading to a maximum utilization \behzad{fraction} of more than 75\% for BRAMs and DSPs. \behzad{ZCU102} is equipped with an 8GB 64-bit DDR-4 off-chip memory. In our implementation, this memory contains input images and CNN parameters. \behzad{It is also} used for interfacing purposes with the host. 

\subsubsection{Undervolting Methodology.}
Unfortunately, there is \behzad{no} voltage scaling standard for FPGAs. Different vendors have their unique voltage management methodologies. Moreover, there are some platforms without voltage scaling capability, such as \behzad{the} Xilinx Zedboard~\cite{zedboard}. Even a single vendor's different devices do not necessarily have the same voltage distribution model. \textcolor{black}{Although this \fff{non-standard} approach of vendors adds some constraints to experimental studies, such as the one conducted in this paper, we believe that, with minor changes, the methodology we explain below for ZCU102 can be applicable to other platforms, as, for instance, we previously studied for on-chip memories of older FPGA generations \cite{salami2018comprehensive}.}  

\begin{figure}[H]
\centering
\includegraphics[width=\linewidth]{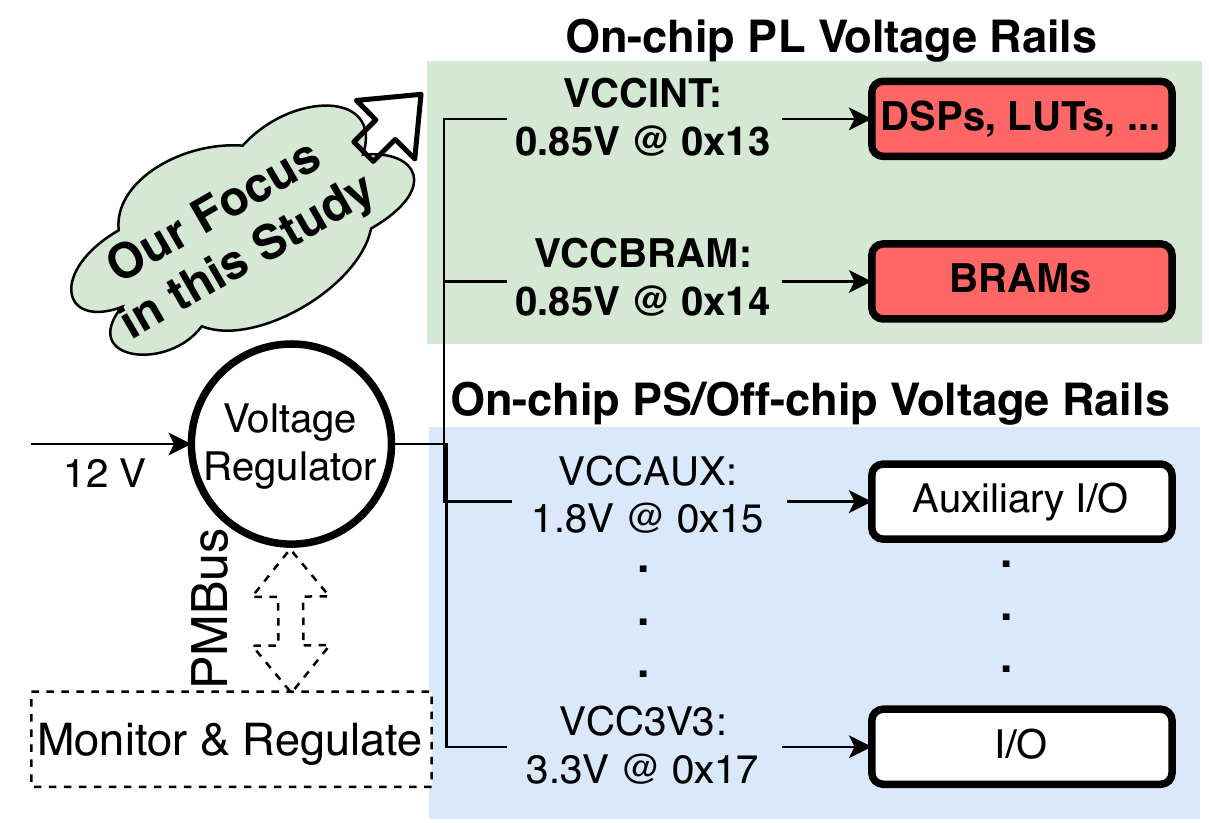}
\caption{Voltage distribution on the Xilinx ZCU102 FPGA, adapted from~\protect\cite{zcu102}.}
\label{fig:overall-voltage} 
\end{figure}

Figure~\ref{fig:overall-voltage}, \behzad{adapted from \cite{zcu102},} depicts the voltage distribution model of ZCU102. \behzad{Here}, the voltage scaling capability is provided using an \behzad{on-board} voltage regulator that can convert an input voltage level of $12V$ into different voltage levels. The voltage level of the output lines, usually called voltage rails, \behzad{is} fully configurable and also addressable using the Power Management Bus (PMBus) \behzad{standard~\cite{pmbus}}. Each voltage rail feeds one or more components of the FPGA platform. ZCU102 is equipped with three voltage regulators, which in total provide 26 voltage rails accessible through the PMBus. In this paper, we focus on on-chip voltage \behzad{rails:} $V_{CCINT}$ and $V_{CCBRAM}$, as shown in \behzad{Figure~\ref{fig:overall-voltage}}. $V_{CCINT}$ is accessible with PMBus address 0x13 and $V_{nom}= 850mV$; it supplies multiple PL components\behzad{, including} DSPs, LUTs, buffers, and routing resources. $V_{CCBRAM}$ is accessible with PMBus \behzad{address} 0x14 and $V_{nom}= 850mV$; it supplies the BRAMs of the PL. \behzad{To} access these voltage rails for monitoring and regulation, we use a PMBus adapter and the provided API~\cite{maxtool}. Using a similar approach and different PMBus commands, we monitor the power consumption of each voltage rail as well as the on-chip temperature.

\section{Experimental Results}
\label{sec:results}
\behzad{We} present and \behzad{analyze our} experimental results \behzad{from reduced-voltage operation on FPGA boards}. 
\behzad{These results are collected} at ambient \behzad{temperature.} Section~\ref{sec:temperature} presents further temperature analysis. \behzad{Each} result presented in this paper is the average of 10 experiments\fff{, in} order to account for any variation between different experiments; although, the variation \behzad{we} observed was negligible.

\subsection{Power Analysis of FPGA-based CNN Accelerators at the Nominal Voltage Level ($V_{nom}$)}
We measure the total on-chip power consumption of the baseline configuration to be \behzad{an average of $12.59W$ for benchmarks}, at \behzad{the} nominal voltage \behzad{level ($V_{nom}$)} and ambient temperature. \behzad{This value} includes the power consumption at on-chip voltage rails, including $V_{CCBRAM}$ and $V_{CCINT}$. We observe that internal FPGA components on \behzad{the $V_{CCINT}$ rail dissipate more than 99.9\% of} this on-chip power. \behzad{We believe} this observation \behzad{is due to} power-efficient BRAM \behzad{designs, using techniques like dynamic power gating~\cite{bramU}}, in modern Ultrascale+ FPGA platforms, including in the studied \behzad{ZCU102 FPGA}. Older generations of Xilinx FPGAs like \behzad{the} 7-series are not equipped with this capability \cite{bram7}\behzad{. Thus}, for \behzad{such} older devices, BRAM power consumption was the main source of \behzad{FPGA} power consumption, \behzad{as shown} in previous studies~\cite{salami2018comprehensive, salami2018fault, salami2019evaluating, ahmed2018automatic}. \behzad{For} the rest of the paper, as we study the power-reliability trade-off, we concentrate on $V_{CCINT}$ \behzad{due to its dominance in FPGA power consumption}.

\subsection{Overall Voltage Behavior}
\begin{figure}[H]
\centering
\includegraphics[width=\linewidth]{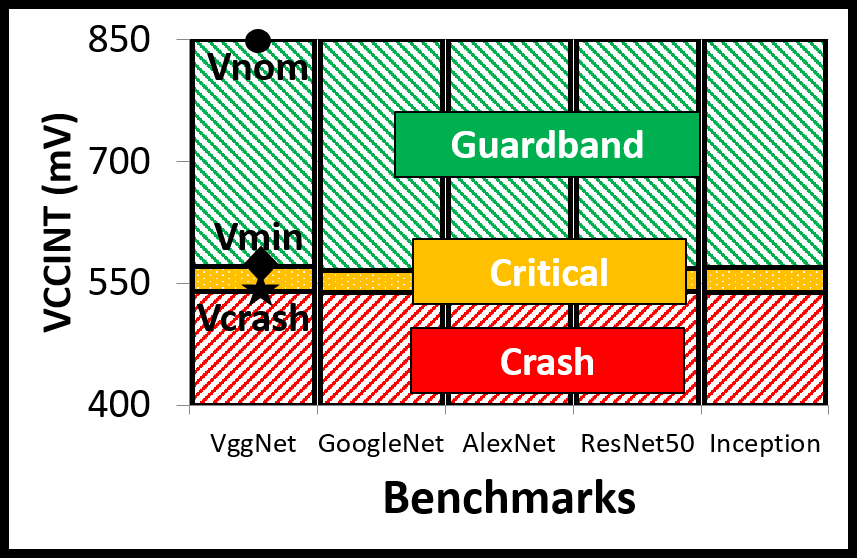}
\caption{Voltage regions with a slight workload-to-workload variation (averaged across three hardware platforms).}
\label{subfig:guardband} 
\end{figure}

\behzad{Our experiments reveal} that a \behzad{large} voltage guardband below $V_{nom}$ exists for $V_{CCINT}$, \behzad{as shown in Figure~\ref{subfig:guardband}} for three hardware platforms and five CNN benchmarks. 
In the voltage guardband region, \behzad{as we reduce supply voltage} there is no performance or reliability degradation, and thus, \behzad{under} normal conditions, eliminating this \behzad{voltage guardband} can lead to significant \behzad{power savings} without any overhead. As Figure~\ref{subfig:guardband} shows, we measure the \behzad{average} guardband \behzad{amount to} be $850mV-570mV= 280mV$, with a slight variation \behzad{across} different \behzad{benchmarks}. In other words, we observe that $V_{min}= 570mV$ (on average) is the minimum safe voltage level of the accelerator, where there is no accuracy \behzad{loss.} \behzad{As we} further \behzad{undervolt} below $V_{min}$, we enter a region called \behzad{the} \textit{critical region} in which the reliability of the hardware and, in turn, the accuracy of the CNN starts to decrease significantly. As Figure~\ref{subfig:guardband} depicts, we measure the \behzad{average} critical voltage region \behzad{size, to be} $570mV-540mv= 30mV$, with a slight variation \behzad{across} different benchmarks. \behzad{As we further undervolt} below $V_{min}$, we reach a \behzad{point at} which the FPGA does not respond to \behzad{requests} and it is not \behzad{functional. This} point is called $V_{crash}$. We \behzad{find that} $V_{crash}= 540mV$ on average, with a slight variation \fff{across different hardware platforms.} 

\begin{figure}[H]
\centering
\includegraphics[width=\linewidth]{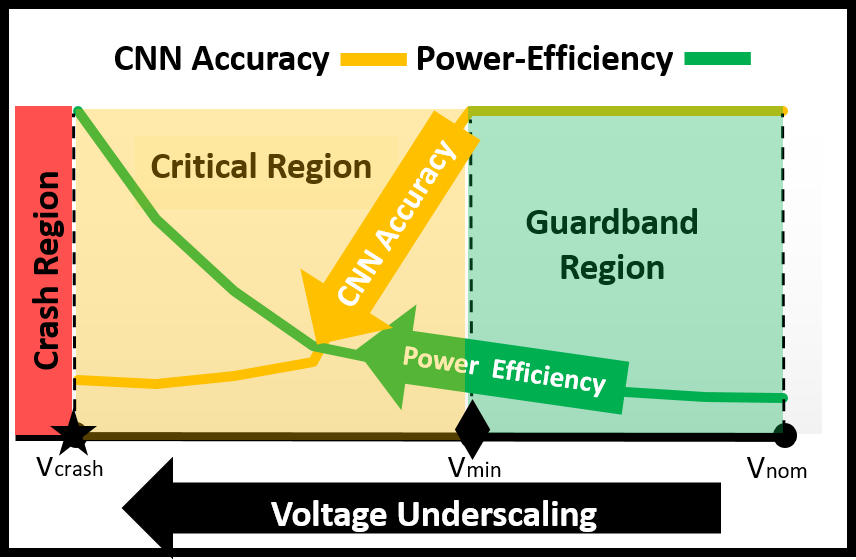}
\caption{Overall voltage behavior observed for $V_{CCINT}$.}
\label{subfig:overallTradeoff} 
\end{figure}

\behzad{Figure~\ref{subfig:overallTradeoff} illustrates} the overall behavior \behzad{we} observe for \behzad{the} power-efficiency and CNN accuracy trade-off on our FPGA-based CNN accelerator. As we perform undervolting, \behzad{the FPGA enters} \fff{the} guardband region\behzad{, where} we observe no reliability degradation (\textit{i.e.,} CNN accuracy loss), and therefore, the power-efficiency comes with no cost. We observe this behavior until we reach the point $V_{min}$, \behzad{\textit{i.e.,}} minimum safe voltage level. With further undervolting, \fff{the FPGA enters} the critical \behzad{region, where} power-efficiency constantly increases\behzad{, but we} start to observe \behzad{fast}-increasing \behzad{CNN} accuracy loss. \behzad{When} we undervolt \behzad{down} to a specific point, called $V_{crash}$, the FPGA becomes non-functional and \behzad{starts to hang}. Sections~\ref{subsec:power} and \ref{subsec:resilience} provide more \fff{details on} the power-reliability trade-off. \textcolor{black}{\behzad{Our} demonstration is on three identical samples of Xilinx ZCU102. However, we believe that the overall voltage behavior, illustrated in Figure~\ref{subfig:overallTradeoff}, is reproducible for other FPGA platforms \behzad{as well}.}

\subsection{Detailed Power-Efficiency Analysis}
\label{subsec:power}
\begin{figure}[H]
\centering
\includegraphics[width=\linewidth]{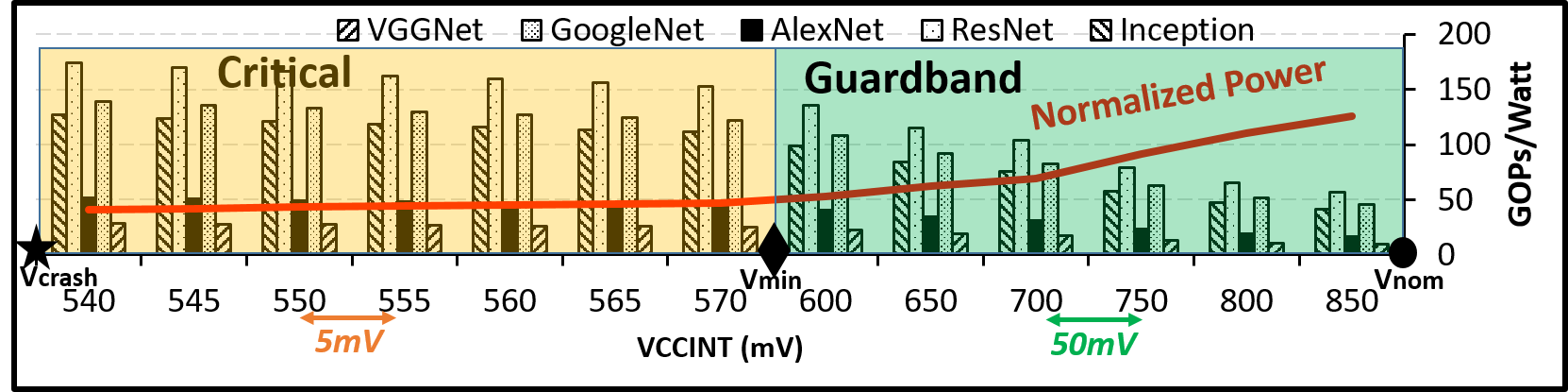}
\caption{Power-efficiency ($GOPs/W$) improvement via undervolting (averaged across three hardware platforms).}
\label{fig:gops} 
\end{figure}

Figure~\ref{fig:gops} presents the power-efficiency experimental results ($GOPs/W$) for five CNN workloads, averaged \behzad{across} three FPGA hardware platforms.  
The power-efficiency gain at $V_{crash}$ is more than 3X \behzad{of that at} nominal voltage level, \textit{i.e.,} $V_{nom}$, for the same design of the given CNN accelerator. 2.6X of the gain in power-efficiency is the result of eliminating the voltage guardband without any \behzad{CNN} accuracy loss. \behzad{43\% further} power-efficiency \behzad{gain} is due to further undervolting in the critical \fff{region,} which has an associated CNN accuracy loss cost. 

\begin{figure*}[hb] 
\centering
\subfloat[VGGNet.]{\includegraphics[width=0.2\textwidth]{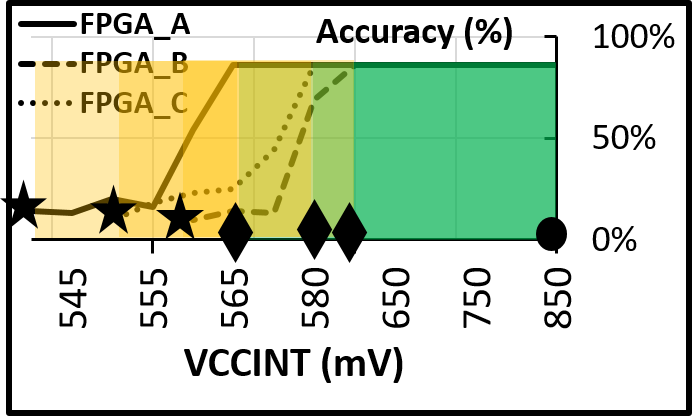}\label{subfig:VGGNet}}
\subfloat[GoogleNet.]{\includegraphics[width=0.2\textwidth]{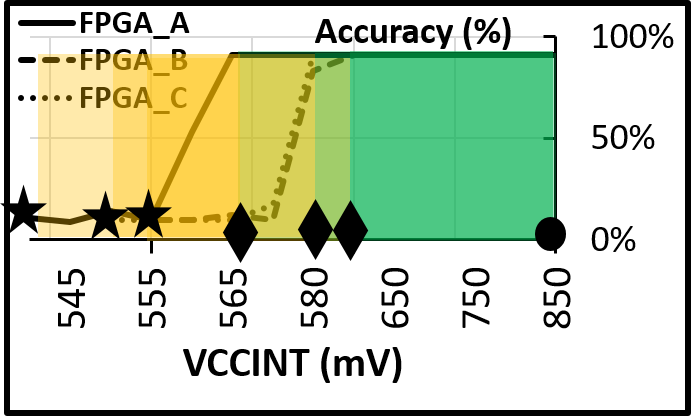}\label{subfig:GoogleNet}}
\subfloat[AlexNet.]{\includegraphics[width=0.2\textwidth]{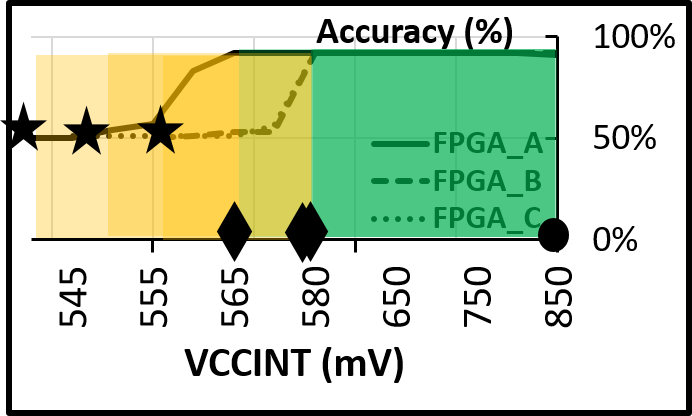}\label{subfig:AlexNet}}
\subfloat[ResNet.]{\includegraphics[width=0.2\textwidth]{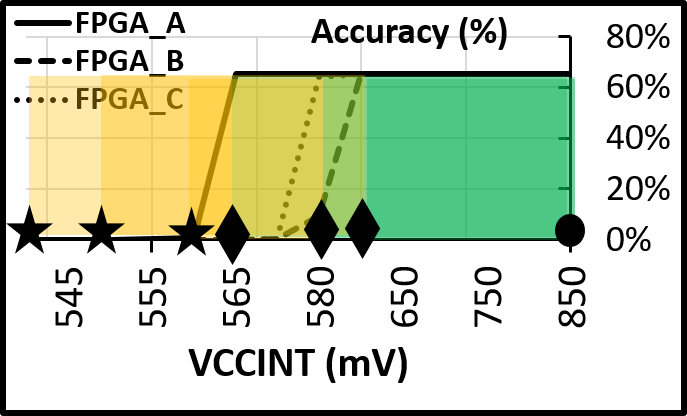}\label{subfig:ResNet}}
\subfloat[Inception.]{\includegraphics[width=0.2\textwidth]{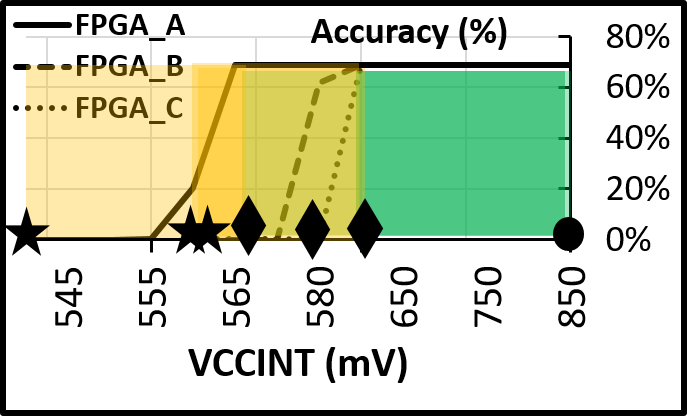}\label{subfig:Inception}}
\caption{Effect of reduced supply voltage on the accuracy of CNN workloads (separately for three hardware platforms).} 
($V_{nom}:\CIRCLE, V_{min}:\blacklozenge, V_{crash}:\bigstar)$
\label{fig:resilience}
\end{figure*}


\behzad{The} power-efficiency \behzad{gain via} undervolting \fff{until $V_{min}$} is not application-dependent, so it is useful for any application mapped onto the same FPGA. However, the reliability overhead in the critical \behzad{region} \fff{below $V_{min}$} is application-dependent due to different vulnerability levels of different applications/workloads. 

\subsection{Detailed Reliability Analysis}
\label{subsec:resilience}
\textcolor{black}{As \behzad{we undervolt} until $V_{min}$, there is no reliability overhead. However, \behzad{as we further undervolt} below $V_{min}$, the reliability of the hardware is significantly affected due to the further increase \behzad{in} datapath delay. The effect of the reliability loss is fully application-dependent due to different \behzad{inherent} resilience levels of different applications. In this paper, we study this effect on several CNN workloads. Figure~\ref{fig:resilience} depicts \behzad{our} experimental results.} As \behzad{shown before}, as we reduce the supply voltage, \behzad{power}-efficiency improves. \behzad{When} we reduce the supply voltage below $V_{min}$, we observe that the accuracy of all benchmarks gradually \behzad{reduces}. With further undervolting, when the supply voltage reaches \behzad{an} average of $V_{crash}=540mV$ \behzad{across} different platforms and benchmarks, the accuracy of the benchmarks drops \behzad{greatly}, and the classifier behaves randomly. Our experiments show that benchmarks with more parameters, \textit{e.g.,} ResNet and Inception are relatively more vulnerable to undervolting faults below $V_{min}$. Also, as seen, there is a variation \behzad{of} $\Delta V_{min}=31mV$ and $\Delta V_{crash}=18mV$ \behzad{across} different \behzad{FPGAs}. This variation can be due to the process variation \behzad{across} different \behzad{FPGAs}.

\section{Frequency Underscaling}
\label{sec:mitigation}
As \behzad{shown} earlier, in the critical voltage \behzad{region} below the guardband, \behzad{CNN} classification accuracy dramatically decreases. In this section, we aim to overcome this \behzad{accuracy loss} by exploiting frequency underscaling. To be more precise, we aim to find a more energy-efficient \behzad{voltage} setting than the undervolted $V_{min}$\behzad{,} which also provides accurate results. \behzad{To this end}, for each supply \behzad{voltage} setting below \behzad{$V_{min}$,} we aim to identify the maximum frequency value $F_{max}$ with which the system does not experience any accuracy loss. When we find this frequency point, we \behzad{evaluate} the energy efficiency of the system. As we underscale the frequency of the system, the performance of the application \behzad{reduces}. Therefore, we \behzad{use} the $GOPs/J$ metric as it accommodates for both performance and energy consumption. 

Table~\ref{table:frequency} summarizes the results of the frequency underscaling in the critical region. \behzad{These} experiments are based on frequency and voltage steps of $25Mhz$ and $5mV$, respectively. The column $V_{CCINT}$ corresponds to the supply voltage of \behzad{a given} setting. The column $F_{max}$ corresponds to the maximum frequency \behzad{at which \fff{there is} no accuracy loss}. The remaining columns: $GOPs$, $Power$, \behzad{$GOPS/W$,} $GOPS/J$ are normalized to the respective values of executing the system in the default setting \behzad{$V_{CCINT}=V_{min}=570mV, F_{max}=333Mhz$} which are the baseline settings \fff{of our accelerator}. Table~\ref{table:frequency} indicates that multiple voltage settings $V_{CCINT}$ map to the same operating Frequency \behzad{$F_{max}$: supply} voltages between $560mV$ to $545mV$ \behzad{require \fff{the} same} frequency of $F_{max}=250Mhz$. \behzad{This is} because the frequency step we use \behzad{is $25Mhz$}. Using smaller steps of frequency can \fff{lead to more spread-out} $F_{max}$ values. 

\begin{table}[H]
\centering
\caption{Evaluation of frequency underscaling to prevent CNN accuracy loss in the critical voltage region (averaged across three hardware platforms). Best result with frequency underscaling in terms of each metric is marked in blue.}
\label{table:frequency}
\begin{tabular}{c|r||rrrr|}
\hline
\multicolumn{1}{|c}{\textbf{\begin{tabular}[c]{@{}c@{}}\small{$V_{CCINT}$}\\ \small{(mV)}\end{tabular}}} & \multicolumn{1}{c||}{\textbf{\begin{tabular}[c]{@{}c@{}}\small{$F_{max}$}\\ \small{(Mhz)}\end{tabular}}} & \multicolumn{1}{c}{\textbf{\begin{tabular}[c]{@{}c@{}}\small{$GOPs$}\\ \small{(Norm)}\end{tabular}}} & \multicolumn{1}{c}{\textbf{\begin{tabular}[c]{@{}c@{}}\small{$Power(W)$}\\ \small{(Norm)}\end{tabular}}} & \multicolumn{1}{c}{\textbf{\begin{tabular}[c]{@{}c@{}}\behzad{\small{$GOPs/W$}}\\ \behzad{\small{(Norm)}}\end{tabular}}}  &
\multicolumn{1}{c|}{\textbf{\begin{tabular}[c]{@{}c@{}}\small{$GOPs/J$}\\ \small{(Norm)}\end{tabular}}} \\ \hline \hline
\multicolumn{1}{|c}{\textbf{570}} & 333 & \textcolor{blue}{\textbf{1.00}} & 1.00 & \behzad{1.00} & \textcolor{blue}{\textbf{1.00}}  \\
\multicolumn{1}{|c}{\textbf{565}} & 300 & 0.94 & 0.97 & \behzad{0.97} & 0.87   \\
\multicolumn{1}{|c}{\textbf{560}} & 250 & 0.83 & 0.84 & \behzad{0.99} & 0.75  \\
\multicolumn{1}{|c}{\textbf{555}} & 250 & 0.83 & 0.78 & \behzad{1.06} & 0.80   \\
\multicolumn{1}{|c}{\textbf{550}} & 250 & 0.83 & 0.75 & \behzad{1.10} & 0.83   \\
\multicolumn{1}{|c}{\textbf{545}} & 250 & 0.83 & 0.74 & \behzad{1.12} & 0.84   \\
\multicolumn{1}{|c}{\textbf{540}} & 200 & 0.70 & \textcolor{blue}{\textbf{0.56}} & \textcolor{blue}{\textbf{1.25}} & 0.75   \\ \hline
\end{tabular}
\end{table}

\behzad{For} all the combinations of $(V_{i}, F_{i})$ \behzad{that} provide error-free results presented \fff{in} Table~\ref{table:frequency} in the critical region, \behzad{power} decreases with \behzad{decreasing} $V_{i} < V_{min}$ and $F_i < F_{max}$. \fff{This is} because we decrease both the supply voltage and the operating frequency. However, at the same time, this leads to decreasing the system performance. Consequently, the best voltage-frequency combination in terms of \emph{energy-efficiency} \fff{($GOPs/J$)} is the one with the highest frequency of $F_{max} = 333Mhz$, which also is our baseline. In other words, it is not worth to underscale the frequency and voltage to find a more energy-efficient optimal point. However, as a trade-off, the design is more \emph{power-efficient} \fff{(\textit{i.e.,} has higher $GOPs/W$) at lower voltage-frequency levels\fff{, up to 25\% at $V_{crash}=540mV$.}}

\section{Combining Undervolting with Architectural CNN Optimization Techniques}
\label{sec:qp}
\label{subsec:architectural}
In this section, we experimentally evaluate undervolting for \behzad{employing the CNN's} quantization and pruning techniques. \behzad{Via} experiments, we observe that \behzad{these} bit reduction techniques can deliver additional power-\behzad{efficiency} gains proportional to the quantization/pruning level. However, applying these techniques can slightly increase the vulnerability of CNNs to undervolting-related faults. \behzad{This section \fff{reports} results \fff{for} VGGNet as we observe similar results for other workloads.}  

\subsection{Quantization}
\behzad{Our} baseline is optimized with INT8 \behzad{precision}. As shown in Table~\ref{table:benchmarks}, this precision does not incur any significant accuracy loss in comparison to baseline models \behzad{that use} floating-point precision. For further analysis of the \behzad{effect} of undervolting with lower precision models, we evaluate INT7, INT6, INT5, and INT4 precisions. \behzad{Using} DNNDK, we observe significant \behzad{accuracy} loss for INT3, INT2, and INT1 when executed at $V_{nom}$\behzad{. Thus}, we do not present them in this paper.

\begin{figure}[H]
\centering
\subfloat[\behzad{CNN} Accuracy.]{\includegraphics[width=0.25\textwidth]{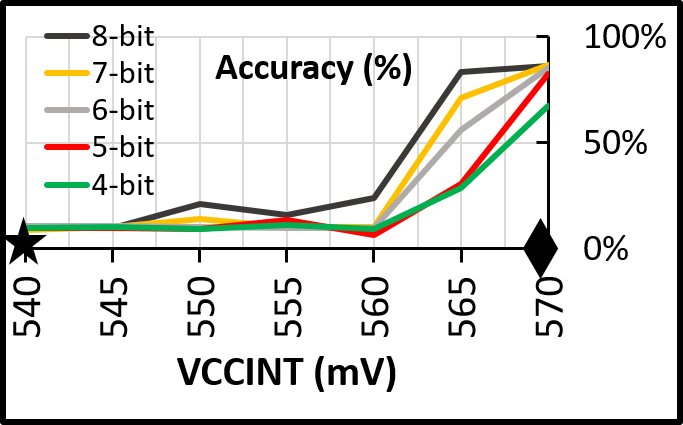}\label{subfig:quantizatonA}}
\subfloat[Power-efficiency ($GOPs/W$).]{\includegraphics[width=0.25\textwidth]{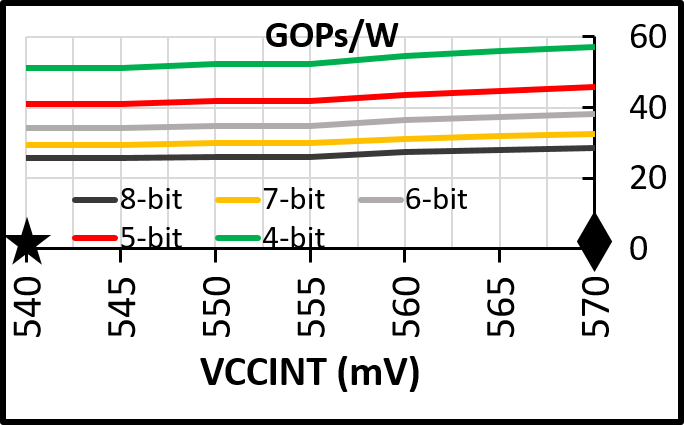}\label{subfig:quantizationP}}
\caption{Effect of undervolting at different quantization levels for VGGNet (averaged across three hardware platforms).}
\label{fig:quantization} 
\end{figure}

Figure~\ref{fig:quantization} shows results of different precisions \behzad{(INT8 to INT4)}. \behzad{We find \fff{that} \textit{i)} when operating at reduced-voltage levels, accuracy loss is relatively high due to lower precision; \textit{ii)} power-efficiency is proportional \fff{to} voltage as well as \fff{quantization} levels.} In conclusion, combining low-\behzad{precision} and low-voltage \behzad{operation} can significantly deliver higher power-efficiency. However, it comes at the cost of accuracy loss. 

\subsection{Pruning}
\begin{figure}[H]
\centering
\subfloat[\behzad{CNN} Accuracy.]{\includegraphics[width=0.25\textwidth]{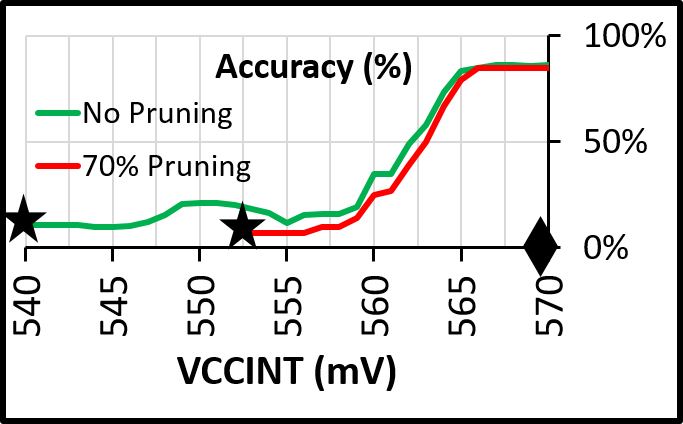}\label{subfig:pruningA}}
\subfloat[Power-efficiency ($GOPs/W$).]{\includegraphics[width=0.25\textwidth]{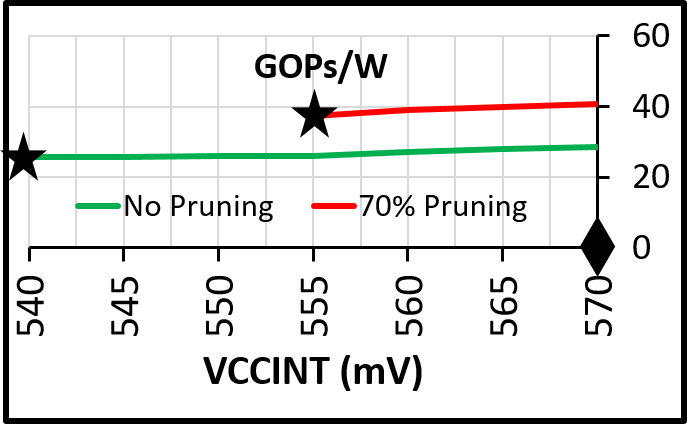}\label{subfig:pruningP}}
\caption{Effect of undervolting on pruned CNN models for VGGNet (averaged across three hardware platforms).}
\label{fig:pruning} 
\end{figure}

Figure~\ref{fig:pruning} shows results \behzad{of pruned and baseline (without any pruning) models}. \behzad{We find that} 
\behzad{undervolting}-related faults have a relatively more significant effect on the pruned model. However, \fff{this} comes with \behzad{higher power}-efficiency of the pruned model, as shown in Figure~\ref{subfig:pruningP}, due \behzad{to fewer} operations \behzad{in} the pruned model. \behzad{With} undervolting, \behzad{power} consumption reduces for both pruned and baseline models, \behzad{at} a similar rate. $V_{crash}$ is different for the pruned model. \behzad{Specifically}, the pruned version demonstrates a higher $V_{crash}$ \behzad{voltage} equal to $555mV$ in contrast to the baseline \behzad{$V_{crash}$} of $540mV$. 

\section{Effect of Environmental Temperature}
\label{sec:temperature}
The power consumption of a modern chip\fff{, including FPGAs,} also depends \fff{on} temperature. \behzad{Temperature} affects \behzad{static} power consumption. As the external temperature increases, the leakage current and, in turn, the leakage-induced static power increases~\cite{Borkar1999,Kaul2009,Kim2003,Huang2011}. \behzad{As technology node size reduces, a large \fff{fraction} of power consumption comes from the static power}. \behzad{Therefore,} temperature has a larger \behzad{effect} on the power consumption of \behzad{denser chips}~\cite{moradi2014side}. On the other \behzad{hand}, \behzad{temperature} can have a considerable \behzad{effect} on \behzad{circuit} latency~\cite{neshatpour2018enhancing,mottaghi2019aging}, \textit{i.e.,}, circuit latency \behzad{\emph{decreases}} as the temperature increases \behzad{in} contemporary technology nodes. \behzad{Therefore,} there are \behzad{fewer} undervolting\behzad{-related} faults at higher temperatures. 

\behzad{To understand the combination of multiple effects mentioned above, we} study the effect of the environmental temperature \behzad{on} the power-reliability trade-off \behzad{of \fff{our} FPGA-based CNN accelerator under reduced-voltage operation}. \behzad{To this end}, we use GoogleNet as a benchmark and undervolt \behzad{$V_{CCINT}$}. We discuss the voltage behavior in both critical and guardband regions \behzad{at} different temperatures ranging from 34$^{\circ}$C to 52$^{\circ}$C degrees. \behzad{To} regulate the FPGA temperature, we control the fan speed using the PMBus interface. \behzad{We also use the same PMBus interface to monitor the on-board live temperature.} By doing so, we can test different ambient \behzad{temperatures} ranging from 34$^{\circ}$C to 52$^{\circ}$C degrees.\footnote{\behzad{[34$^{\circ}$C, 52$^{\circ}$C]} is the temperature range that we could generate using the fan speed. Experimenting \behzad{with} wider temperature ranges \behzad{requires} more facilities, which were not available \behzad{to us}.} 

\subsection{Temperature Effect on Power Consumption}
\label{ssec:voltNetTempPower}
\begin{figure}[H]
\centering
\includegraphics[width=\linewidth]{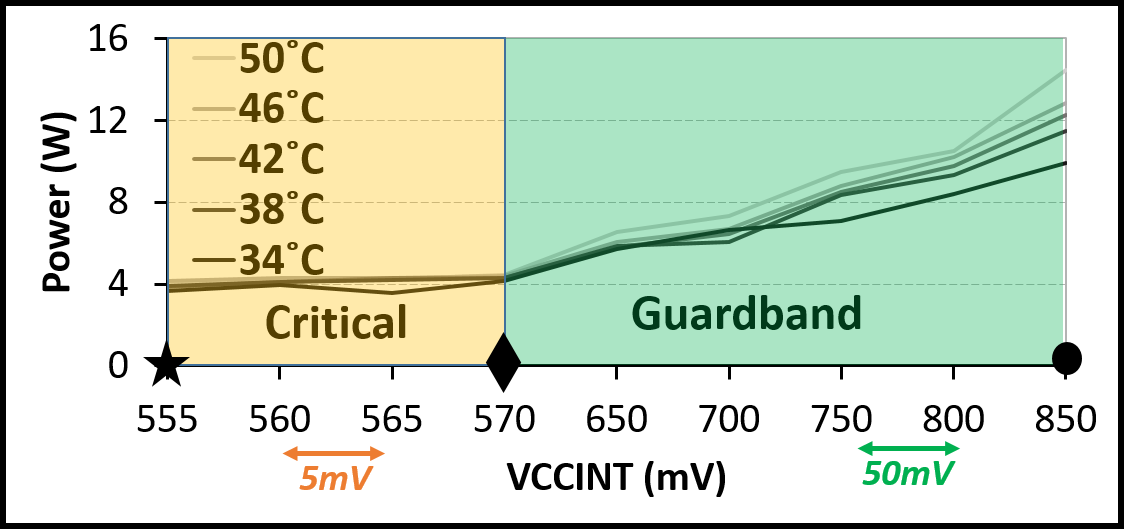}
\caption{Power consumption of our reduced-voltage CNN accelerator at temperature range of [34$^{\circ}$C, 52$^{\circ}$C], shown for GoogleNet (averaged across three hardware platforms).}
\label{subfig:temperatureP} 
\end{figure}

Figure~\ref{subfig:temperatureP} depicts the power consumption of \fff{our} \behzad{CNN} accelerator when executing GoogleNet \behzad{with} different $V_{CCINT}$ values \behzad{at different temperatures.} \behzad{Clearly,} temperature has a direct \behzad{effect on} power consumption. \behzad{As} temperature increases, power consumption proportionally increases. This is due to \behzad{increase in} static power when the chip heats up. \behzad{Dynamic} power consumption is \behzad{also} affected by \behzad{temperature}, but \behzad{this effect} is almost negligible. \behzad{Importantly,} we observe \behzad{that} \behzad{the effect of} temperature \behzad{on} power consumption \behzad{reduces} for lower voltages. For \behzad{example} power change from 34$^{\circ}$C to 52$^{\circ}$C \behzad{are 0.46\% and 0.15\%, respectively} at $V_{CCINT}=850mV$ and $V_{CCINT}=650mV$.

\subsection{Temperature Effect on Reliability}
\begin{figure}[H]
\centering
\includegraphics[width=\linewidth]{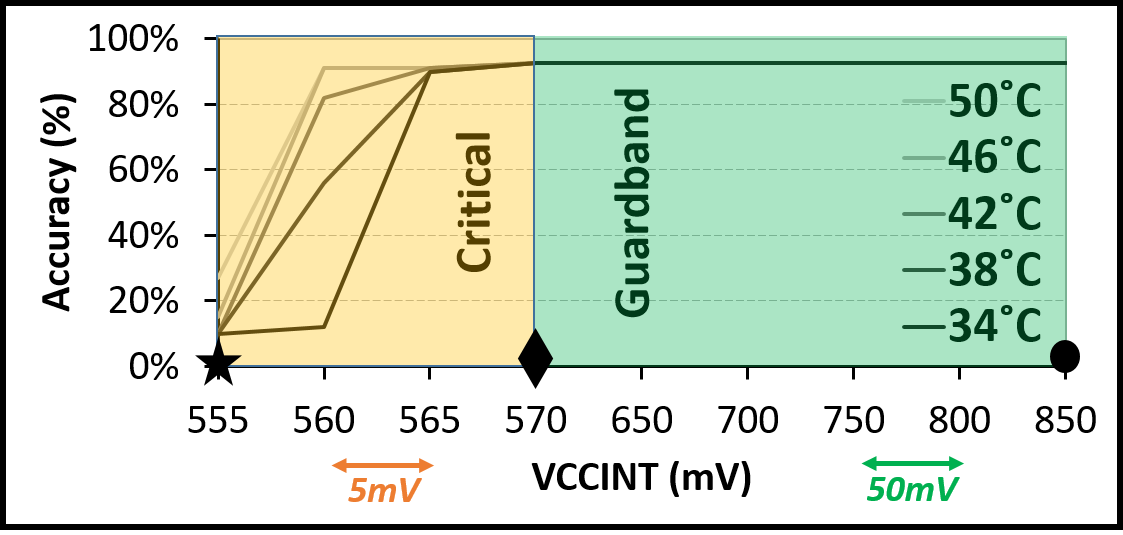}
\caption{Accuracy of our reduced-voltage CNN accelerator at temperature range of [34$^{\circ}$C, 52$^{\circ}$C], shown for GoogleNet (averaged across three hardware platforms).}
\label{subfig:temperatureA} 
\end{figure}

\behzad{Figure~\ref{subfig:temperatureA} shows} the \behzad{effect} \behzad{of temperature on} the accuracy of \behzad{our reduced-voltage CNN} accelerator. Our experiment demonstrates that \behzad{\textit{i)}} there is no \fff{noticeable} change in the \behzad{size} of \fff{the} guardband and critical regions, and \behzad{\textit{ii)}} 
\behzad{higher} temperature at a particular voltage level \behzad{leads to higher} CNN accuracy. \behzad{This is because} at higher temperatures, there are \behzad{fewer} undervolting related errors due to decreased circuit latency, an artifact due \behzad{to} \fff{the} Inverse Thermal Dependence \behzad{(ITD) property} \fff{of} contemporary technology nodes~\cite{neshatpour2018enhancing,Uht2004}. 

\subsection{Discussion}
In our setup, \behzad{considering the power-reliability trade-off discussed,} the optimal setting is at Temp=50$^{\circ}$ and $V_{CCINT}=565mV$, \behzad{\textit{i.e.,}} the minimum voltage level at which there is almost no accuracy loss due to the healing effect of high temperature. \behzad{However}, the disadvantage of operating at higher temperatures \behzad{is} the overall decrease in lifetime reliability. Below, we summarize our findings \behzad{on} temperature effects. 
\begin{itemize}
 \item There is a negligible change in the value of $V_{min}=570mV$ \behzad{across temperatures}, and thus, there is no significant change \behzad{in the} guardband region. However, the system crashes relatively earlier over temperature variation. We expect, though, that when the system \behzad{undergoes a} wider temperature \behzad{range,} there will be \behzad{a more noticeable change} in the $V_{min}$ and $V_{crash}$. 
\item At any specific voltage \behzad{point} in \fff{either region}, \behzad{power} consumption directly increases as temperature increases, mainly due to the direct relation \behzad{of static power consumption} and temperature.
\item The \behzad{effect of temperature on power} consumption is significantly less at lower voltage levels, due to the relatively lower contribution \behzad{of static} power \behzad{to total power consumption}. 
\item In the critical voltage region and at any specific voltage level, higher temperature leads to \behzad{higher CNN accuracy}. The power cost of the higher temperature in the critical voltage region is \behzad{relatively low}. 
\end{itemize}

Consequently, a lower voltage can be applied at higher temperatures without causing significant accuracy loss \fff{at a small power cost}.


\section{Related Work}
\label{sec:related}
\behzad{To our knowledge, this paper provides the first study evaluating the effect of reduced-voltage operation in FPGA-based CNN accelerators. In this section, we review related works \fff{on} \textit{i)} undervolting, \textit{ii)} power-efficient CNNs, and \textit{iii)} reliability of CNNs.}

\subsection{Undervolting}
Supply voltage underscaling below the nominal level is \fff{an} effective approach to improve the power-efficiency \behzad{of digital} circuits. There are two different approaches \behzad{to studying undervolting: simulation or real \fff{experiments}}.

\subsubsection{Simulation Studies.}
\behzad{This approach simulates hardware to study undervolting. It is convenient for early-stage studies as \behzad{it} does not require large engineering \fff{effort}. However, this approach lacks the information of real hardware, and thus, \behzad{validation of} results \behzad{is} the main concern.} Most of the existing simulation-based studies are for \behzad{CPUs}~\cite{roelke2017pre,yalcin2016exploring,swaminathan2017bravo,8416495} and specifically for \behzad{CPU components} \fff{such as caches~\cite{alameldeen2010adaptive,wilkerson2008trading,wilkerson2010reducing,yalcin2014exploiting,chishti2009improving}} and branch predictors~\cite{chatzidimitriou2019assessing}. \behzad{There are also studies for} ASIC CNN accelerators~\cite{reagen2016minerva,zhang2018thundervolt,andri2017yodann}. \behzad{Following this approach, studies on FPGA}-based designs are either \behzad{fully in} simulation~\cite{mottaghi2019aging} or emulation of FPGA netlists \behzad{on simulation} frameworks~\cite{khaleghi2019fpga,salamat2019workload}.

\subsubsection{Experimental Studies on Real Hardware.}
\label{subsec:relatedUndervolting}
Evaluating \behzad{undervolting} on real hardware is another approach that has recently been considered for multiple devices~\cite{gizopoulos2019modern, george2020exceeding}. \behzad{Doing so requires} relatively more engineering effort \behzad{as well as considering} physical constraints, such \behzad{as} non-standard device- and vendor-dependent voltage distribution \behzad{models. Yet,} the results produced are accurate and can be directly used in real-world applications. 

\fff{Undervolting} of real hardware \behzad{is studied for various system components,} such as CPUs~\cite{bacha2013dynamic,papadimitriou2017harnessing,papadimitriou2017voltage,kaliorakis2018statistical,bertran2014voltage}, GPUs~\cite{zou2018voltage, leng2015gpu,leng2015safe}, ASICs~\cite{chandramoorthy2019resilient,pandey2019greentpu,kim2018matic}, \behzad{DRAMs~\cite{chang2017understanding, chang2018voltron,koppula2019eden},} and \behzad{Flash disks~\cite{cai2013threshold, cai2015read, cai2017error}}. \behzad{These studies focus} on voltage guardband analysis, fault characterization, and fault mitigation. \behzad{Undervolting on real} FPGAs is not thoroughly \behzad{investigated.} \behzad{Very recent works} on \fff{FPGA} undervolting are either accompanied with \behzad{frequency} underscaling~\cite{ahmed2018automatic,shen2019fast} that can diminish \behzad{performance, or are} limited \behzad{to} BRAMs~\cite{salami2018demo,salami2018comprehensive, salami2018fault, salami2019evaluating,salami2018aggressive}. This paper\behzad{, for the first time,} extends \behzad{real FPGA undervolting studies to} multiple on-chip components of modern FPGA fabrics and evaluates it in-detail on the power-accuracy trade-off of CNN applications. 

\subsection{Power-efficient CNNs}
\behzad{Many works aim to improve CNN power-efficiency by optimizing \fff{the} CNN architecture as well as the underlying hardware.} \fff{In this paper, to achieve significant power-efficiency, we combine our hardware-level FPGA undervolting technique with architectural CNN optimization techniques, including quantization and pruning.}  
\subsubsection{Architectural Techniques.}
This approach aims to reduce the parameter size of a CNN. The methods of this approach are independent of the underlying hardware, and in theory, they can be applied to any hardware\behzad{, including} hardware accelerators. The most common techniques are \fff{quantization}~\cite{zhou2017incremental, han2016eie,zhu2019configurable}, \fff{pruning}~\cite{molchanov2016pruning,yazdani2018dark,han2015learning}, batching~\cite{shen2017escher}, loop unrolling~\cite{ma2017optimizing}, and memory compression~\cite{deng2018permdnn,kim2015compression}. Among \behzad{these}, \behzad{quantization and pruning} have shown significant efficiency without \fff{significantly} compromising the CNN accuracy; hence, we \behzad{focus on} them in our experiments. 

\subsubsection{Hardware-level Techniques.}
\behzad{An orthogonal approach to reducing CNN power is to optimize the underlying hardware.} \fff{To} this \behzad{end}, \behzad{since} traditional processor-based architectures are power-hungry and not suitable for CNNs, exploiting a dedicated hardware is the first approach. \behzad{Further} power \behzad{savings are possible with} low-level techniques, such as \behzad{undervolting.} 

\begin{itemize}
    \item Hardware Accelerators: Data-flow execution models using GPUs~\cite{khorasani2018register,hill2017deftnn}, FPGAs~\cite{sharma2016high,suda2016throughput,xiao2017exploring,ma2018optimizing,li2019rnn} and ASICs~\cite{jouppi2017datacenter,andri2016yodann,chen2014diannao,wang2019bit} are more efficient choices \behzad{for CNNs than} traditional CPUs. Among \behzad{these}, FPGAs are more flexible compared to ASICs and more efficient than GPUs. \behzad{Efficient} exploitation of the underlying hardware is fundamental \behzad{for power-efficiency}, using \behzad{techniques like resource partitioning~\cite{shen2017maximizing}} and \behzad{reuse~\cite{zhang2015optimizing,riera2018computation}. Our work uses} an industrial \behzad{tool~\cite{dnndk}} that inherently exploits these techniques. 
    \item Undervolting: \behzad{Undervolting \fff{has been shown to provide} significant power-efficiency benefit for CNNs \fff{when applied to} SRAMs~\cite{chandramoorthy2019resilient},  DRAMs~\cite{koppula2019eden}, ASICs~\cite{moons201714, zhang2018thundervolt, yang2017sram,chandramoorthy2019resilient,kim2018matic}, and heterogeneous systems~\cite{ cristal2018legato2,cristal2018legato1,salami2020legato}}.   
\end{itemize}

\subsection{Reliability of CNNs}
Although CNNs are inherently resilient to some error rate \behzad{in data or underlying hardware}, \behzad{high enough error} rates can \behzad{cause} significant accuracy \behzad{loss}. \behzad{Error sources} can be harsh environments, process manufacturing defects, undervolting, ionizing particles, \behzad{noise in data,} among others. Hence, \behzad{CNN reliability is an active research area.} \behzad{Existing} studies \behzad{are based on fault} injection or \behzad{real errors.}

\subsubsection{Simulation-based Fault Injection.}
\behzad{These studies} inject randomly-generated faults \behzad{into} CNNs, \behzad{but they do not consider undervolting}~\cite{jha2019ml,salami2018resilience,reagen2018ares,li2017understanding,leng2020asymmetric, givaki2019resilience,jha2019kayotee,liu2017fault}. This approach provides an opportunity \behzad{for} comprehensive fault characterization \behzad{of} CNNs, such as the sensitivity of different layers, different location of faults, among others. However, \behzad{these works} do not consider \behzad{faults in real hardware\fff{, which potentially can lead to inaccurate analysis}}.

\subsubsection{Faults in Real Hardware.}
In real-world applications, such as IoT, airspace, and driver-less cars, CNNs can potentially experience different \behzad{types} of faults. \behzad{Various works evaluate CNN reliability on faulty real hardware, \textit{e.g.,} soft errors~\cite{libano2018selective, libano2020understanding,trindade2019assessment,brewer2019impact} and undervolting in ASICs~\cite{li2019chip,chandramoorthy2019resilient, whatmough201714, whatmough2018dnn, lee201916}. This approach requires significant engineering effort but can result in relatively more accurate results. None of these works study CNN reliability on undervolted FPGAs.} 

\section{Summary and Future Work}
\label{sec:conclusion}
In this paper, we \behzad{experimentally} evaluated the effects of supply voltage underscaling below the nominal level on real FPGA-based CNN accelerators. We \behzad{showed that we could improve} the power-efficiency of \behzad{such} accelerators by more than 3X \behzad{via} \fff{undervolting}. 2.6X of the power-efficiency improvement comes from eliminating the voltage guardband (without compromising CNN accuracy), while the remaining 43\% improvement comes from undervolting further below the guardband \behzad{(which comes \fff{with} CNN accuracy loss). \fff{We conclude that undervolting can significantly improve the power-efficiency of FPGA-based neural network accelerators.}}

As future work, we aim to develop \behzad{\textit{i)}} fault mitigation techniques for low-voltage regions even when the design operates at the maximum frequency ($F_{max}$), \behzad{\textit{ii)}} dynamic voltage adjustment techniques \behzad{considering} temperature, accuracy, power consumption, and performance trade-off. \textcolor{black}{We also aim to expand our experiments in \fff{hardware}\behzad{, by} evaluating more FPGAs\behzad{, as} well as in software\behzad{, by} repeating experiments on other CNN platforms like DNNWeaver~\cite{sharma2016high}. Finally, we \behzad{believe it is promising to} study potential security issues of FPGA-based CNN accelerators under reduced supply voltage levels.} 

\section*{Acknowledgments}
We thank \behzad{the} anonymous \behzad{DSN2020} reviewers for their feedback and comments, as well as Dr. Long Wang, who helped us with \behzad{shepherding}. Also, we thank Dr. Konstantinos Parasyris for his in-depth review of the first version of this paper. The work done for this paper was partially supported by a HiPEAC Collaboration Grant funded by the H2020 HiPEAC Project under grant agreement \behzad{No.} 779656. \behzad{The} research leading to these results has received funding from the European Union’s Horizon 2020 Programme under the LEGaTO Project (www.legato-project.eu), grant agreement \behzad{No.} 780681. This work is supported in part by funding from the SRC and gifts from Intel, Microsoft and VMware to Onur Mutlu.
\bibliographystyle{plain}
\bibliography{references}

\end{document}